\PassOptionsToPackage{dvipsnames,table}{xcolor}
\documentclass[table]{article} 
\usepackage{iclr2024_conference,times}

\usepackage{amsmath,amsfonts,bm}

\def\eqref#1{equation~\ref{#1}}

\def\1{\bm{1}}

\DeclareMathAlphabet{\mathsfit}{\encodingdefault}{\sfdefault}{m}{sl}
\SetMathAlphabet{\mathsfit}{bold}{\encodingdefault}{\sfdefault}{bx}{n}

\usepackage[utf8]{inputenc} 
\usepackage[T1]{fontenc}    
\usepackage{hyperref}       
\usepackage{url}            
\usepackage{booktabs}       
\usepackage{amsfonts}       
\usepackage{nicefrac}       
\usepackage{microtype}      
\usepackage{times}
\usepackage{epsfig}
\usepackage{graphicx}
\usepackage{amsmath}
\usepackage{amssymb}
\usepackage{pdfrender}

\usepackage{hyperref}
\usepackage{url}
\hypersetup{colorlinks}
\usepackage{mathtools}
\usepackage{algorithm}
\usepackage{soul}
\usepackage{tc}
\usepackage{enumerate}
\usepackage{subcaption}
\usepackage[noend]{algpseudocode}
\usepackage{makecell}
\usepackage{comment}
\usepackage{caption}
\usepackage{subcaption}
\usepackage{enumitem}
\usepackage{amsthm}
\usepackage{arydshln}
\usepackage{bm}
\usepackage{siunitx,tabularx,ragged2e,booktabs}
\usepackage{wrapfig}
\usepackage{float}
\usepackage{booktabs}
\usepackage{pifont}
\usepackage{epstopdf}
\usepackage{blindtext}
\usepackage{fancyvrb}
\usepackage{cleveref}
\usepackage[table]{xcolor}

\usepackage[outline]{contour}
\contourlength{0.25pt}
\contournumber{20}

\newcommand{\xmark}{\ding{55}}%

\newcommand{\imagenet}{\text{ImageNet}}

\newcommand{\algname}{automated search-space generation neural architecture search}

\newcommand{\algacro}{ASGNAS{}}
\newcommand{\dhspg}{\text{DHSPG}}

\newcommand{\hhspg}{\text{H2SPG}}

\newcommand{\cifar}{\text{CIFAR10}}

\newcommand{\fashionmnist}{\text{Fashion-MNIST}}
\newcommand{\stl}{\text{STL-10}}
\newcommand{\svnh}{\text{SVNH}}

\newcommand{\demosupnet}{\text{DemoNet}}

\newcommand{\stackedunets}{\text{StackedUnets}}
\newcommand{\superresnet}{\text{SuperResNet}}
\newcommand{\darts}{\text{DARTS}}

\newcommand{\regnet}{\text{RegNet}}

\newcommand{\ie}{\textit{i.e.}}
\newcommand{\eg}{\textit{e.g.}}

\newcommand{\xcheckmark}{\checkmark\kern-1.1ex\raisebox{.7ex}{\rotatebox[origin=c]{125}{--}}}

\usepackage{framed}
\usepackage{tikz}
\usetikzlibrary{arrows}
\usepackage{setspace}
\tikzset{
  treenode/.style = {align=center, inner sep=0pt, text centered,
    font=\sffamily},
  arn_n/.style = {treenode, circle, white, font=\sffamily\bfseries, draw=black,
    fill=black, text width=1.5em},
  arn_r/.style = {treenode, circle, red, draw=red,
    text width=1.5em, very thick},
  arn_x/.style = {treenode, rectangle, draw=black,
    minimum width=0.5em, minimum height=0.5em}
}

\fvset{frame=single,framesep=1mm,fontfamily=courier,fontsize=\scriptsize,numbers=left,framerule=.3mm,numbersep=1mm,commandchars=\\\{\}}

\title{Automated Search-Space Generation Neural Architecture Search}

\author{Tianyi Chen, \ Luming Liang, \ Tianyu Ding, \ Ilya Zharkov\\
\small{Microsoft}\\
\small{Redmond, WA 98052, USA}\\
\footnotesize{\texttt{\{tiachen,lulian,tianyuding,zharkov\}@microsoft.com}} \\
}

\iclrfinalcopy
\begin{document}

\maketitle

\begin{abstract}
To search an optimal sub-network within a general deep neural network (DNN), existing neural architecture search (NAS) methods typically rely on handcrafting a search space beforehand. Such requirements make it challenging to extend them onto general scenarios without significant human expertise and manual intervention. 
To overcome the limitations, we propose \textbf{A}utomated \textbf{S}earch-Space \textbf{G}eneration \textbf{N}eural \textbf{A}rchitecture \textbf{S}earch (\textbf{\algacro{}}), perhaps the first automated system to train general DNNs that cover all candidate connections and operations and produce high-performing sub-networks in the one shot manner. Technologically, \algacro{} delivers three noticeable contributions to minimize human efforts: \textit{(i)} automated search space generation for general DNNs; \textit{(ii)} a Hierarchical Half-Space Projected Gradient (\hhspg{}) that leverages the hierarchy and dependency within generated search space to ensure the network validity during optimization,  and reliably produces a solution with both high performance and hierarchical group sparsity; and \textit{(iii)} automated sub-network construction upon the \hhspg{} solution. Numerically, we demonstrate the effectiveness of \algacro{} on a variety of general DNNs, including RegNet, StackedUnets, SuperResNet, and DARTS, over benchmark datasets such as \cifar{},~\fashionmnist{},~\imagenet{},~\stl{}, and~\svnh{}. The sub-networks computed by \algacro{} achieve competitive even superior performance compared to the starting full DNNs and other state-of-the-arts. 
\end{abstract}

\section{Introduction}\label{sec.introduction}

Deep neural networks (DNNs) have achieved remarkable success in various fields, which success is highly dependent on their sophisticated underlying architectures~\citep{lecun2015deep,goodfellow2016deep}. To design effective DNN architectures, human expertise have handcrafted numerous popular DNNs such as ResNet~\citep{he2016deep} and transformer~\citep{NIPS2017_3f5ee243}. However, such human efforts may not be scalable enough to meet the increasing demands for customizing DNNs for diverse tasks. To address this issue, Neural Architecture Search (NAS) has emerged to automate the network creations and reduce the need for human expertise~\citep{elsken2018efficient}.  

In the realm of NAS studies, discovering the optimal sub-network within a general DNN that covers all candidate connections and operations stands as a pivotal topic. Gradient-based methods~\citep{liu2018darts,yang2020ista,xu2019pc,chen2021progressive} are perhaps the most popular for the discovery because of their efficiency. Such methods parameterize operation candidates via introducing auxiliary architecture variables with weight sharing, then search a (sub)optimal sub-network via formulating and solving a multi-level optimization problem.

Despite the advancements in gradient-based NAS methods, their usage is still limited due to certain inconvenience. In particular, their automation relies on manually determining the search space for a pre-specified DNN beforehand, and requires the manual introduction of auxiliary architecture variables onto the prescribed search space. To extend these methods onto other DNNs, the end-users still need to manually construct the search pool, then incorporate the auxiliary architecture variables along with building the whole complicated multi-level optimization training pipeline. The whole process necessitates significant domain-knowledge and engineering efforts, thereby being inconvenient and time-consuming for users. Therefore, it is natural to ask whether we could reach an

\textit{\textbf{Objective.} Given a general DNN, automatically generate its search space, train it once, and construct a sub-network that achieves a dramatically compact architecture and high performance.}

\begin{figure}
\centering
\includegraphics[width=0.95\linewidth]{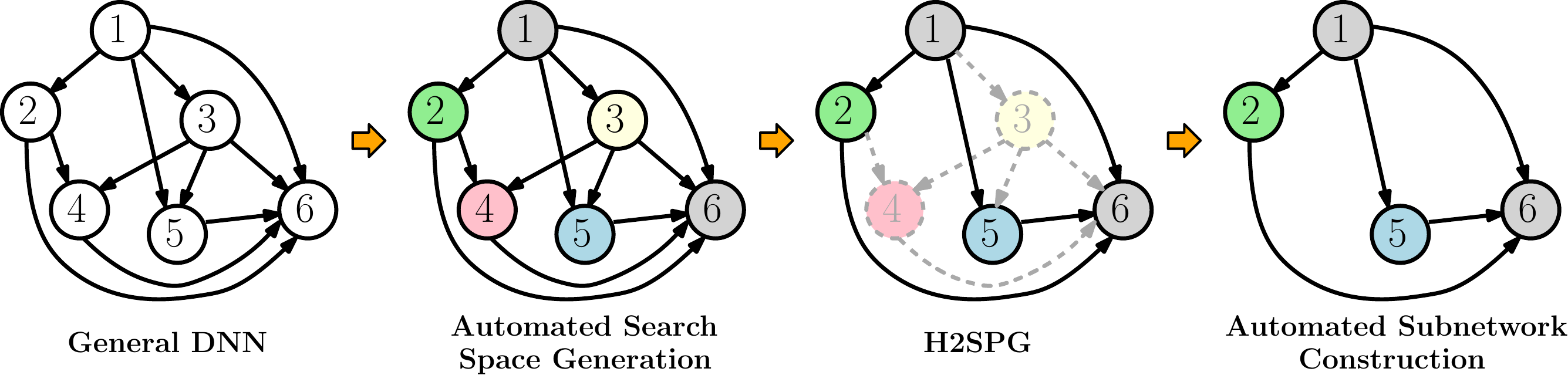}
\caption{Overview of \algacro{}. Given a general DNN, \algacro{} first automatically generates a search space, then employs \hhspg{} to identify redundant removal structures and train the important counterparts to high-performance, finally constructs a compact and high-performing sub-network.}
\label{fig:overview_asgnas}
\vspace{-3mm}
\end{figure}

Achieving the objective is severely challenging in terms of both engineering developments and algorithmic designs, consequently not achieved yet by the existing works to the best of our knowledge. We now build \algname~(\algacro{}) that first reaches the objective. Given a DNN that covers all operation and connection candidates, \algacro{} automatically generates a search space, trains and identifies redundant structures, then builds a sub-network that achieves both high performance and compactness, as shown in Figure~\ref{fig:overview_asgnas}. The whole procedure can be automatically proceeded, dramatically reduce the human efforts, and fit general DNNs and applications. Our main contributions can be summarized as follows. 

\begin{itemize}[leftmargin=*]
\item \textbf{Automated Search Space Generation and Sub-Network Construction.} We propose a novel graph algorithm to automatically exploit the architecture given a general DNN, then analyze the hierarchy and dependency across different operators to form a search space. The established search space consists of the structures that could be removed without interrupting the functionality of the remaining DNN. We further propose a novel graph algorithm to automatically construct a sub-network upon the starting DNN parameterized as the subsequent \hhspg{} solution.

\item \textbf{Hierarchical Half-Space Projected Gradient (\hhspg{}).} We propose a novel \hhspg{}, perhaps the first optimizer, that solves a hierarchical structured sparsity problem for general DNN applications. \hhspg{} computes a solution of both high performance and desired sparsity level. Compared to other sparse optimizers, \hhspg{} conducts a dedicated hierarchical search phase over the generated search space to ensures the validness of the constructed sub-network.  

\item \textbf{Experimental Results.} We demonstrate the effectiveness of \algacro{} on extensive DNNs including \regnet{}, \stackedunets{}, \superresnet{} and \darts{}, over benchmark datasets including \cifar{}, \fashionmnist{}, \imagenet{}, \stl{}, and \svnh{}. \algacro{} is the first framework that could automatically deliver compact sub-networks upon general DNNs to the best of our knowledge. Meanwhile the sub-networks exhibit competitive even superior performance to the full networks.

\end{itemize}

\section{Related Work}\label{sec.related_work}

\paragraph{Neural Architecture Search (NAS).} 

Early NAS works utilized reinforcement learning and evolution techniques to search for high-quality architectures~\citep{zoph2016neural,pham2018efficient,zoph2018learning}, while they were computationally expensive. Later on, differentiable (gradient-based) methods were introduced to accelerate the search process. These methods start with a over-parameterized DNN covering all possible connection and operation candidates, and parameterize them with auxiliary architecture variables. They establish a multi-level optimization problem that alternatingly updates the architecture and network variables until convergence~\citep{liu2018darts,chen2019progressive,xu2019pc,yang2020ista,hosseini2022saliency}.
However, these methods require a significant amount of \textbf{handcraftness} from users in advance to \textbf{manually} establish the search space, introduce additional architecture variables, and build the multi-level training pipeline. The sub-network construction is also network-specific and not flexible. All requirements necessitate remarkable domain-knowledge and expertise, making it difficult to extend to broader scenarios. 
\vspace{-2mm}

\paragraph{Hierarchical Structured Sparsity Optimization.} We formulate the underlying 
optimization problem of \algacro{} as a hierarchical
structured sparsity problem. Its solution possesses high group sparsity indicating redundant structures and obeys specified hierarchy. There exist deterministic optimizers solving such problems via introducing latent variables~\citep{zhao2009composite}, while are impractical for stochastic DNN tasks. Meanwhile, stochastic optimizers rarely study such problem. In fact, popular stochastic sparse optimizers such as HSPG~\citep{chen2020half,chen2023otov2}, proximal methods~\citep{xiao2014proximal} and ADMM~\citep{lin2019toward} overlook the hierarchy constraint. Incorporating them into \algacro{} typically delivers invalid sub-networks. Therefore, we propose \hhspg{} that considers the graph topology to ensure the validity of produced sub-networks.

\section{\algacro{}}\label{sec.otov3}
\vspace{-1mm}

\algacro{} is an automated one-shot system designed to train a general DNN and subsequently construct a sub-network. The resulting sub-network is not only high-performing but also has a remarkably compact architecture, making it well-suited for various deployment environments. The entire process of \algacro{} significantly reduces the necessity for human intervention and is compatible with a wide range of DNNs and applications. As outlined in Algorithm~\ref{alg:main.outline}, \algacro{} takes a starting DNN $\mathcal{M}$, explores its trace graph, examines the inherent hierarchy, and autonomously constructs a search space (Section~\ref{sec.auto_search_space_construction}). Based on the hierarchy presented within the search space, corresponding trainable variables are segregated into a series of groups, adhering to structural constraints. Subsequently, a hierarchical structured sparsity optimization problem is articulated and addressed through a novel approach—Hierarchical Half-Space Projected Gradient (H2SPG) (Section~\ref{sec.dhspg_plus}). \hhspg{} takes into account the hierarchy embedded within the generated search space and calculates a solution that achieves both high performance and the desired sparsity level. Ultimately, a compact sub-network $\mathcal{M}^*$ is constructed by eliminating the structures associated with identified redundant structures and their dependent modules (Section~\ref{sec.auto_sub_net_construction}). \vspace{-1mm}


\begin{algorithm}[h!]
	\caption{Outline of \algacro{}.}
	\label{alg:main.outline}
	\begin{algorithmic}[1]
		\State \textbf{Input:} A general DNN $\mathcal{M}$ to be trained and 
        searched (no need to be pretrained). 
		\State \textbf{Automated Search Space Generation.} Analyze the trace graph of $\mathcal{M}$, generate a search space, and partition the trainable parameters into a set of groups obeying the hierarchy of search space. 
		\State \textbf{Train by \hhspg{}.} Seek a high-performing solution with hierarchical group sparsity.
		\State \textbf{Automated Sub-Network Construction.} Construct a sub-network $\mathcal{M}^*$ upon \hhspg{} solution.
		\State \textbf{Output:} Constructed sub-network $\mathcal{M}^*$. (Post fine-tuning is optional).
 \end{algorithmic}
\end{algorithm}
\vspace{-2mm}

\subsection{Automated Search Space Generation}\label{sec.auto_search_space_construction} 

\begin{figure}[t]
    \centering
    \begin{subfigure}{\linewidth}
    	\centering
    	\includegraphics[width=\linewidth]{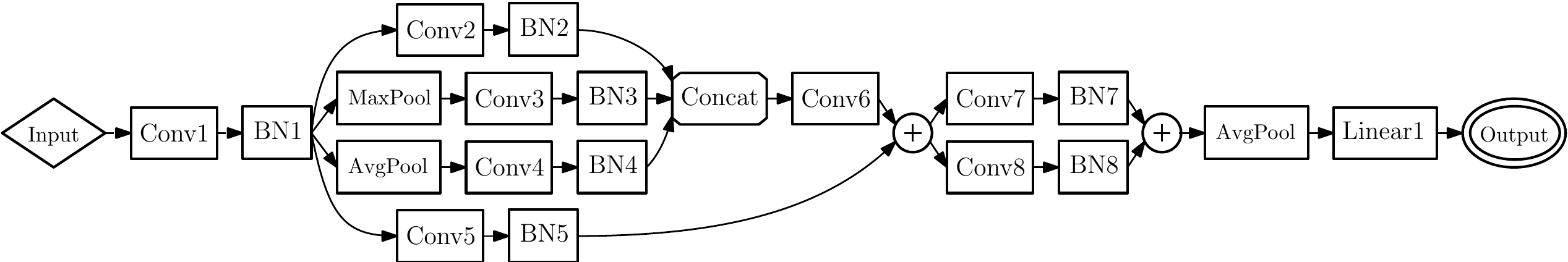}
    	\caption{A demo DNN (\demosupnet) to be trained and searched.}\label{fig:demo_super_network}
    	\vspace{1mm}
    	\label{fig:1a}		
    \end{subfigure}
    \begin{subfigure}{\linewidth}
    	\centering
    	\includegraphics[width=1.0\linewidth]{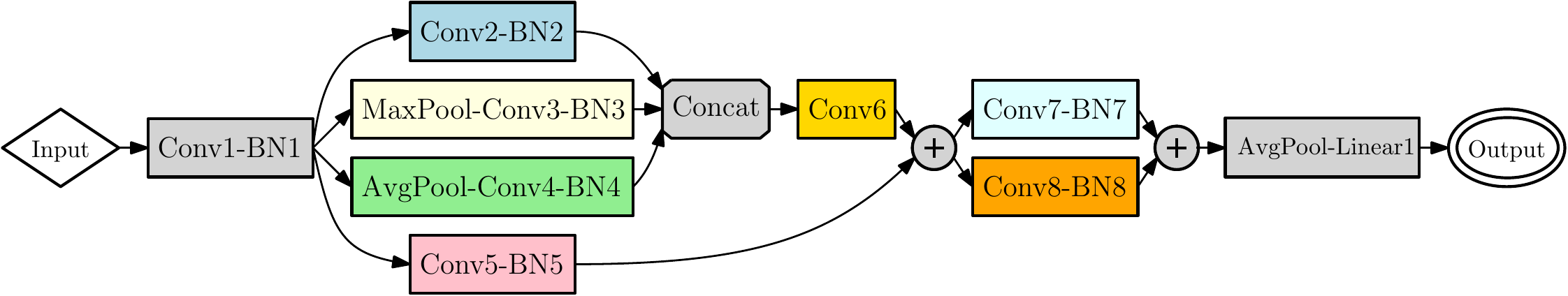}
    	\caption{Segment graph.}\label{fig:dependancy_graph}
    	\vspace{1mm}
    	\label{fig:1a}		
    \end{subfigure}
    \begin{subfigure}{\linewidth}
    	\centering
    	\includegraphics[width=\linewidth]{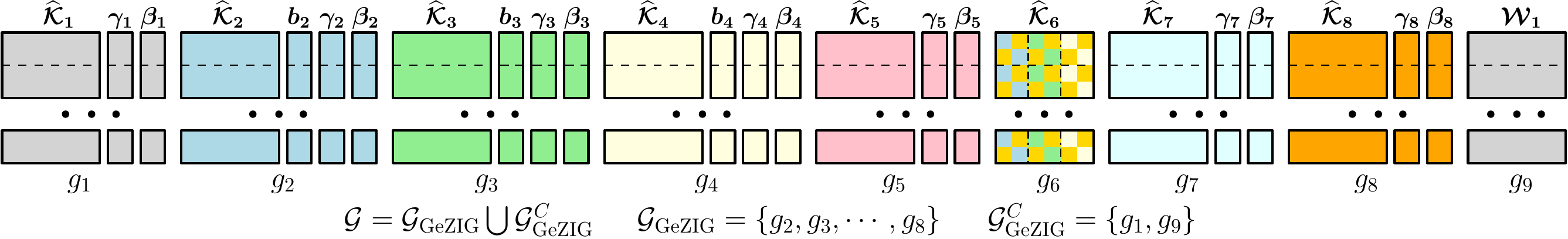}
    	\caption{Trainable variable partition.}\label{fig:ge_zigs}		
    \end{subfigure}
    \caption{Automated Search Space Generation. (a) The \demosupnet{} to be trained and searched; (b) the constructed segment graph; and (c) the trainable variable partition, where $\mathcal{G}_s$ represents the groups corresponding to removal structures. $\bm{\widehat{\mathcal{K}}}_i$ and $\bm{b}_i$ are the flatten filter matrix and bias vector for \texttt{Conv-i}, respectively. $\bm{\gamma}_i$ and $\bm{\beta}_i$ are the weight and bias vectors for \texttt{BN-i}. $\bm{{\mathcal{W}}}_i$ is the weight matrix for \texttt{Linear-i}. The columns of $\bm{\widehat{\mathcal{K}}}_6$ are marked in accordance to its incoming segments.}
    \label{fig:automatic_search_space_construction} 
    \vspace{-3mm}
\end{figure}

The initial step of \algacro{} is to automatically generate a search space for a general DNN, which definition is varying upon distinct NAS scenarios (see more in Appendix~\ref{appendix.more_related_work}).  In our context, the search space is {defined} as the set of structures that can be omitted from the given DNN while ensuring that the remaining network continues to function normally. We refer to such structures as the removal structures of DNNs. Consequently, the generation of the search space is formulated as the discovery of these removal structures. This process poses significant challenges, encompassing both engineering developments and algorithmic designs. These challenges arise due to the intricate architecture of DNNs, the distinct roles of operators, and a scarcity of sufficient public APIs. To address these challenges and accomplish our goal, we have developed a dedicated graph algorithm stated as Algorithm~\ref{alg:main.zig_partition}. The generation of search space involves two main phases. The first phase explores the trace graph of the DNN $\mathcal{M}$ and establishes a segment graph $(\mathcal{V}_s, \mathcal{E}_s)$. The second phase leverages the affiliations inside the segment graph to find out removal structures, then partitions their trainable variables to a set of groups. For intuitive illustrations, we elaborate the algorithm through a small but complex demo DNN depicted in Figure~\ref{fig:demo_super_network}.

{
\begin{algorithm}[h]
	\caption{Automated Search Space Generation.}
	\label{alg:main.zig_partition}
	\begin{algorithmic}[1]
		\State \textbf{Input:} A super-network $\mathcal{M}$ to be trained and searched. 
        \State \textit{\textbf{Segment graph construction.}}
        \State Construct the trace graph $(\mathcal{V}, \mathcal{E})$ of $\mathcal{M}$.\label{line:vertices_edges_trace_graph}
        \State Initialize an empty graph $(\mathcal{V}_{s},\mathcal{E}_{s})$.
        \State Initialize queue $\mathcal{Q}\gets \{\mathcal{S}(v): v\in\mathcal{V}\text{ is adjacent to the input of trace graph}\}$.\label{line.initialize_queue}
        \While{$\mathcal{Q}\neq \emptyset$}
            \State Dequeue the head segment $\mathcal{S}$ from $\mathcal{Q}$.
            \State Grow $\mathcal{S}$ in the depth-first manner till meet either joint vertex or multi-outgoing vertex $\hat{v}$.\label{line.meet_endpoint_vertex}
            \State Add segments into $\mathcal{V}_{s}$ and connections into $\mathcal{E}_{s}$.
            \State Enqueue new segments into the tail of $\mathcal{Q}$ if $\hat{v}$ has outgoing vertices.\label{line.enqueue_new_elements}   
        \EndWhile
        \State \textit{\textbf{Discovery of removal structures.}}
        \State Get the incoming vertices $\widehat{\mathcal{V}}$ for joint vertices in the $(\mathcal{V}_s, \mathcal{E}_s)$.\label{line.incoming_vertices_of_joint}
        \State Group the trainable variables in the vertex $v\in\widehat{\mathcal{V}}$ as $g_v$.\label{line:group_variables_vertex_zig}
        \State Form $\mathcal{G}_s$ as the union of the above groups, \ie, $\mathcal{G}_s\gets\{g_v: v\in \widehat{\mathcal{V}}\}$.
        \label{line.form_zigs}
        \State Form $\mathcal{G}_s^C$ as the union of the trainable variables in the remaining vertices. 
	\State \textbf{Return} trainable variable partition $\mathcal{G}=\mathcal{G}_s\cup \mathcal{G}_s^C$ and segment graph $(\mathcal{V}_{s},\mathcal{E}_{s})$.
	\end{algorithmic}
\end{algorithm}
}


\textbf{Segment Graph Construction.} 
Given a general DNN $\mathcal{M}$, we first construct its trace graph $(\mathcal{V}, \mathcal{E})$ displayed as Figure~\ref{fig:demo_super_network} (line~\ref{line:vertices_edges_trace_graph} in Algorithm~\ref{alg:main.zig_partition}), where $\mathcal{V}$ represents the set of vertices (operations) and $\mathcal{E}$ represents the connections among them. We particularly refer vertices as joint vertices if they aggregate multiple inputs into a single output, \eg, \texttt{Add} and \texttt{Concat}.


{We then analyze the trace graph $(\mathcal{V}, \mathcal{E})$ to create a segment graph $(\mathcal{V}_s, \mathcal{E}_s)$, wherein each vertex in $\mathcal{V}_s$ serves as a potential  removal structure candidate. To proceed, we use a queue container $\mathcal{Q}$ to track the candidates (line \ref{line.initialize_queue} of Algorithm~\ref{alg:main.zig_partition}). The initial elements of this queue are the vertices that are directly adjacent to the input of $\mathcal{M}$, such as \texttt{Conv1}. We then traverse the graph in the breadth-first manner, iteratively growing each element (segment) $\mathcal{S}$ in the queue until a valid removal structure candidate is formed. The growth of each candidate follows the depth-first search to recursively expand $\mathcal{S}$ until the current vertices are considered as endpoints. The endpoint vertex is determined by whether it is a joint vertex or has multiple outgoing vertices, as indicated in line~\ref{line.meet_endpoint_vertex} of Algorithm \ref{alg:main.zig_partition}. Intuitively, a joint vertex has multiple inputs, which means that the DNN may be still valid after removing the current segment. This suggests that the current segment may be removable. On the other hand, a vertex with multiple outgoing neighbors implies that removing the current segment may cause some of its children to miss the input tensor. For instance, removing \texttt{Conv1-BN1} would cause \texttt{Conv2}, \texttt{MaxPool} and \texttt{AvgPool} to become invalid due to the absence of input in Figure~\ref{fig:demo_super_network}. Therefore, it is risky to remove such candidates. Once the segment $\mathcal{S}$ has been grown, new candidates are initialized as the outgoing vertices of the endpoint and added into the container $\mathcal{Q}$ (line~\ref{line.enqueue_new_elements} in Algorithm~\ref{alg:main.zig_partition}). Such procedure is repeated until the end of traversal and returns a segment graph $(\mathcal{V}_s, \mathcal{E}_s)$ in  Figure~\ref{fig:dependancy_graph}.}

\textbf{Discovery of Removal Structures.} We proceed to identify the removal structures in $(\mathcal{V}_s, \mathcal{E}_s)$ to generate the search space. The qualified instances are the vertices in $\mathcal{V}_s$ that have trainable variables and all of their outgoing vertices are joint vertices. This is because a joint vertex has multiple inputs and remains valid even after removing some of its incoming structures, as indicated in line~\ref{line.incoming_vertices_of_joint} in Algorithm~\ref{alg:main.zig_partition}. 
Consequently, their trainable variables are grouped together into $\mathcal{G}_s$ (line \ref{line:group_variables_vertex_zig}-\ref{line.form_zigs} in Algorithm~\ref{alg:main.zig_partition} and Figure~\ref{fig:ge_zigs}). The remaining vertices are considered as either unremovable or belonging to a large removal structure, which trainable variables are grouped into the $\mathcal{G}_s^C$ (the complementary to $\mathcal{G}_s$). As a result, for the given DNN $\mathcal{M}$, all its trainable variables are encompassed by the union $\mathcal{G}=\mathcal{G}_s\cup \mathcal{G}_s^C$, and the corresponding removal structures in $\mathcal{G}_s$ constitute one search space of $\mathcal{M}$. 
\vspace{-2mm}

\subsection{Hierarchical Half-Space Projected Gradient (\hhspg)}\label{sec.dhspg_plus}
\vspace{-1mm}

Given a general DNN $\mathcal{M}$ and its group partition $\mathcal{G}=\mathcal{G}_s\cup \mathcal{G}_s^C$, the next is to jointly search for a valid sub-network $\mathcal{M}^*$ that exhibits the most significant performance and train it to high performance. Searching a sub-network is equivalent to identifying the redundant structures in the search space $\mathcal{G}_s$ to be further removed and ensures the remaining network still valid. Training the sub-network becomes optimizing over the remaining important groups in $\mathcal{G}$ to achieve high performance. We formulate a hierarchical structured sparsity problem to accomplish both tasks simultaneously as follows.
\begin{equation}\label{prob.main}
\minimize{\bm{x}\in \mathbb{R}^n}\ f(\bm{x}),\ \ \text{s.t.} \ \text{Cardinality}(\mathcal{G}^{0})=K,\ \text{and}\  (\mathcal{V}_s/\mathcal{V}_{\mathcal{G}^{0}}, \mathcal{E}_s/\mathcal{E}_{\mathcal{G}^{0}})\ \text{is valid},
\end{equation}
\begin{wrapfigure}{r}{0.6\textwidth}
\vspace{-8mm}
\begin{minipage}{\linewidth}
\begin{algorithm}[H]
\caption{Hierarchical Half-Space Projected Gradient}
\label{alg:main.dhspg}
\begin{algorithmic}[1]
\State \textbf{Input:} initial variable $\bm{x}_0\in\mathbb{R}^n$, initial learning rate $\alpha_0$, target group sparsity $K$, segment graph $(\mathcal{V}_{s}, \mathcal{E}_s)$ and group partition $\mathcal{G}=\mathcal{G}_s\cup \mathcal{G}_s^C$.
\State \textit{\textbf{Hierarchical Search Phase.}}
\State Initialize redundant removal structures $\mathcal{G}_r\gets \emptyset$. 
\State Initialize remaining segment graph $(\widehat{\mathcal{V}}, \widehat{\mathcal{E}})\gets (\mathcal{V}_s, \mathcal{E}_s)$.
\State Calculate the saliency score via modular proxy for each $g\in \mathcal{G}_s$ and sort them.  \label{h2spg.salience_score_calculation}
\For {$g\in \mathcal{G}_s$ ordered by saliency scores ascendingly}\label{line.check_start}
    \State Find the vertex $v_g$ for $g$ and the adjacent edges $\mathcal{E}_g$. \label{line.find_vertice_edges}
    \If {$(\widehat{\mathcal{V}}/\{v_g\}, \widehat{\mathcal{E}}/ \mathcal{E}_g)$ is valid and $|\mathcal{G}_r|< K$}\label{line.check_valid_remain_graph}
        \State Update $\mathcal{G}_r\gets \mathcal{G}_r\cup \{g\}$.
        \State Update $(\widehat{\mathcal{V}},\widehat{\mathcal{E}})\gets(\widehat{\mathcal{V}}/\{v_g\}, \widehat{\mathcal{E}}/ \mathcal{E}_g)$. \label{line.check_end}
    \EndIf
\EndFor
\State \textit{\textbf{Hybrid Training Phase.}}
\For {$t=0, 1,\cdots, $}
	\State Compute gradient estimate $\Grad f(\bm{x}_t)$ or its variant.\label{line.compute_gradient_estimate}
	\State Update $[\bm{x}_{t+1}]_{\mathcal{G}_{r}^C}$ as 
	$[\bm{x}_t-\alpha_t \Grad f(\bm{x}_t)]_{\mathcal{G}_{r}^C}.
	$\label{line.update_important_groups}
	\State Perform Half-Space projection over $[\bm{x}_{t}]_{\mathcal{G}_{r}}$.
	\label{line:half_space_projection}
\EndFor
\State \textbf{Return} the final or the best iterate as $\bm{x}^*_{\hhspg{}}$.
\end{algorithmic}
\end{algorithm}
\end{minipage}
\vspace{-4mm}
\end{wrapfigure}
where $f$ is the prescribed loss function, $\mathcal{G}^{=0}:=\{g\in\mathcal{G}_s| [\bm{x}]_g=0\}$ is the set of zero groups in $\mathcal{G}_s$, which cardinality measures its size. $K$ is the target hierarchical group sparsity, indicating the number of removal structures that should be identified as redundant. The trainable variables in redundant removal structures are projected onto zero, while the trainable variables in important structures are preserved as non-zero and optimized for high performance. A larger $K$ dictates a higher sparsity level that produces a more compact sub-network with fewer FLOPs and parameters.  $(\mathcal{V}_s/\mathcal{V}_{\mathcal{G}^{0}}, \mathcal{E}_s/\mathcal{E}_{\mathcal{G}^{0}})$ refers to the graph removing vertices and edges corresponding to zero groups $\mathcal{G}^0$. It being valid requires the zero groups distributed obeying the hierarchy of the segment graph to ensure the resulting sub-network functions correctly.

\begin{figure}[t]
\centering
\includegraphics[width=\linewidth]{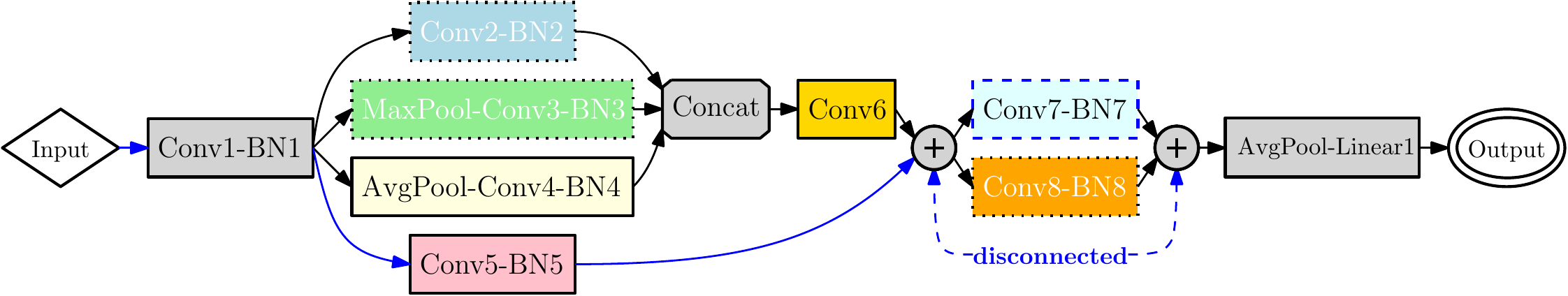}
\caption{Check validness of redundant candidates. Target group sparsity $K=3$. \texttt{Conv7-BN7} has smaller  salience score than \texttt{Conv2-BN2}. Dotted vertices are marked as redundant candidates.} \label{fig:check_validaness_group_sparsity} 
\vspace{-5mm}
\end{figure}

Problem~(\ref{prob.main}) is difficult to solve due to the non-differential and non-convex sparsity constraint and the graph validity constraint. Existing optimizers such as HSPG~\citep{chen2020half,dai2023adaptive} and proximal methods~\citep{deleu2021structured} overlook the architecture evolution and hierarchy during the sparsity exploration, which is crucial to (\ref{prob.main}). In fact, they are mainly applied for orthogonal and distinct pruning tasks, where the connections and operations are preserved yet become slimmer. Consequently, employing them onto (\ref{prob.main}) usually produces invalid sub-networks. 
\vspace{-2mm}

\paragraph{Outline of \hhspg{}.} To effectively solve problem (\ref{prob.main}), we propose a novel \hhspg{} to consider the hierarchy and ensure the validness of graph architecture after removing redundant vertices during the optimization process. 
To the best of our knowledge, \hhspg{} is the first the optimizer that successfully solves such hierarchical structured sparsity problem (\ref{prob.main}), which outline is stated in Algorithm~\ref{alg:main.dhspg}. 

H2SPG is a hybrid multi-phase optimizer, distinguished by its dedicated designs catering to the hierarchical constraint, positioning it significantly apart from its non-hierarchical counterparts within the HSPG sparse optimizer family~\citep{chen2020half,chen2023otov2,dai2023adaptive}. Initially, \hhspg{} categorizes groups of variables into important and potentially redundant segments through a hierarchical search phase. Subsequently, it applies specified updating mechanisms to different segments to achieve a solution with both desired hierarchical group sparsity and high performance via a hybrid training phase.  The hierarchical search phase considers the topology of segment graph $(\mathcal{V}_s, \mathcal{E}_s)$ to ensure the validness of the resulting sub-network. Vanilla stochastic gradient descent (SGD) or its variant such as Adam~\citep{kingma2014adam} optimizes the important segments to achieve the high performance. Half-space gradient descent~\citep{chen2020half} identifies redundant groups among the candidates and projects them onto zero without sacrificing the objective function to the largest extent.

\textbf{Hierarchical Search Phase.}  To proceed, \hhspg{} first computes the saliency scores for each removal structures in the generated search space (line~\ref{h2spg.salience_score_calculation} in Algorithm~\ref{alg:main.dhspg}). The saliency score calculation is modular to varying proxies, \eg, gradient-based proxies or training-free zero-shot proxies \citep{ming_zennas_iccv2021,chen2021neural,li2023zico}
We by default proceed the gradient-based proxy due to its flexibility on general applications and DNNs. In particular,  we first warm up all variables by conducting SGD or its variants. During the warm-up, a saliance score of each group $g\in\mathcal{G}_s$ is computed and exponentially averaged. Smaller salience score indicates the group exhibits less prediction power, thus may be redundant. By default, we consider both the cosine similarity between negative gradient $-[\Grad f(\bm{x})]_g$ and the projection direction $-[\bm{x}]_g$ as well as the average variable magnitude. The former one measures the approximate degradation onto the objective function over the projection direction, and the latter one measures the distance to the origin.

The next is to form a set of redundant removal structure candidates $\mathcal{G}_r$ and ensures the validity of remaining DNN after erasing these candidates (line~\ref{line.check_start}-\ref{line.check_end} in Algorithm~\ref{alg:main.dhspg}). To proceed, we iterate each group in $\mathcal{G}_s$ in the ascending order of salience scores. A remaining graph $(\widehat{\mathcal{V}}, \widehat{\mathcal{E}})$ is constructed by iteratively removing the vertex of each group along with the corresponding adjacent edges from $(\mathcal{V}_s, \mathcal{E}_s)$. The sanity check verifies whether the graph $(\widehat{\mathcal{V}}, \widehat{\mathcal{E}})$ is still connected after the erasion. If so, the variable group for the current vertex is added into $\mathcal{G}_r$; otherwise, the subsequent group is turned into considerations. As illustrated in Figure~\ref{fig:check_validaness_group_sparsity}, though \texttt{Conv7-BN7} has a smaller salience score than \texttt{Conv2-BN2}, \texttt{Conv2-BN2} is marked as potentially redundant but not \texttt{Conv7-BN7} since there is no path connecting the input and the output of the graph after removing \texttt{Conv7-BN7}. This mechanism largely guarantees that even if all redundant candidates are erased, the resulting sub-network is still functioning as normal. The complementary groups with higher redundancy scores are marked as important groups and form $\mathcal{G}_{r}^C:=\mathcal{G}/\mathcal{G}_r$.

\textbf{Hybrid Training Phase.} \hhspg{}{} then engages into the hybrid training phase to produce desired group sparsity over $\mathcal{G}_r$ and optimize over $\mathcal{G}_r^C$ for pursuing excellent performance till the convergence. This phase mainly follows \citep{chen2023otov2,dai2023adaptive}, and is briefly described for completeness. In general, for the important groups of variables in $\mathcal{G}_r^C$, the vanilla SGD or its variant is employed to minimize the objective function (line~\ref{line.compute_gradient_estimate}-\ref{line.update_important_groups} in Algorithm~\ref{alg:main.dhspg}). For redundant group candidates in $\mathcal{G}_{r}$, a Half-Space projection step is proceeded to progressively yield sparsity without sacrificing the objective function and preserving the knowledge over redundant groups to the largest extent. At the end, a high-performing solution $\bm{x}_\hhspg^*$ with desired hierarchical group sparsity is returned.

\vspace{-1mm}
\subsection{Automated Sub-Network Construction.}\label{sec.auto_sub_net_construction}
\vspace{-1mm}

\begin{figure}[t]
    \centering
    \begin{subfigure}{\linewidth}
    	\centering
    	\includegraphics[width=\linewidth]{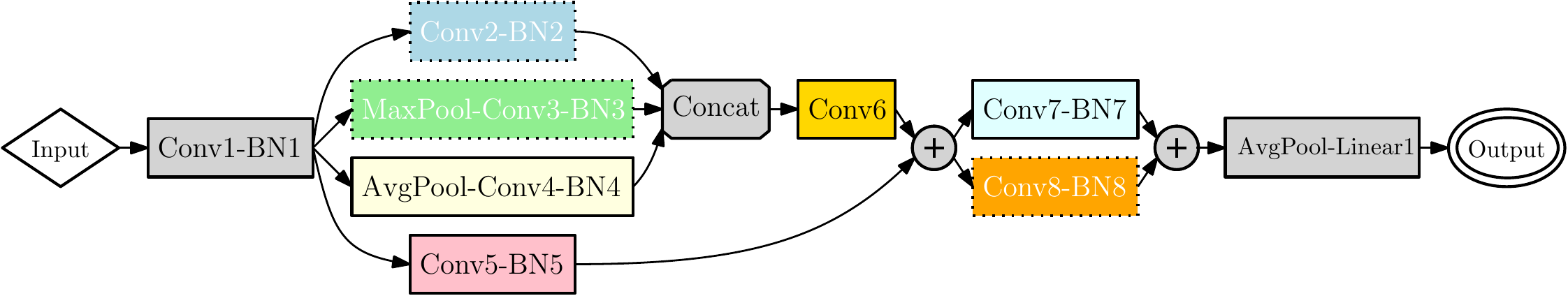}
    	\caption{Identified redundant structures.}\label{fig:redundant_structures}
    	\label{fig:1a}		
    \end{subfigure}
    \begin{subfigure}{0.45\linewidth}
    	\centering
    	\includegraphics[width=\linewidth]{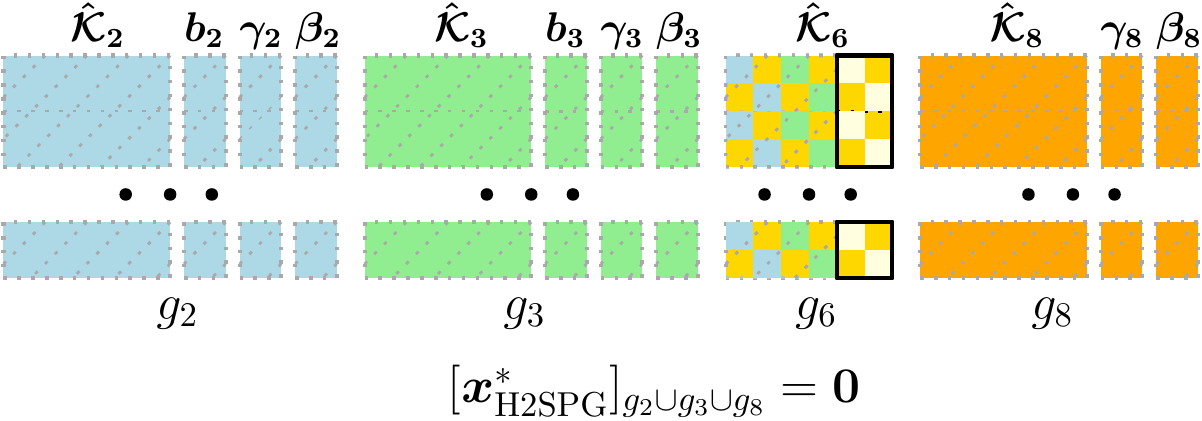}
    	\caption{Redundant removal structures.}\label{fig:zero_groups}
    	\label{fig:1a}		
    \end{subfigure}
    \begin{subfigure}{0.54\linewidth}
    	\centering
    	\includegraphics[width=\linewidth]{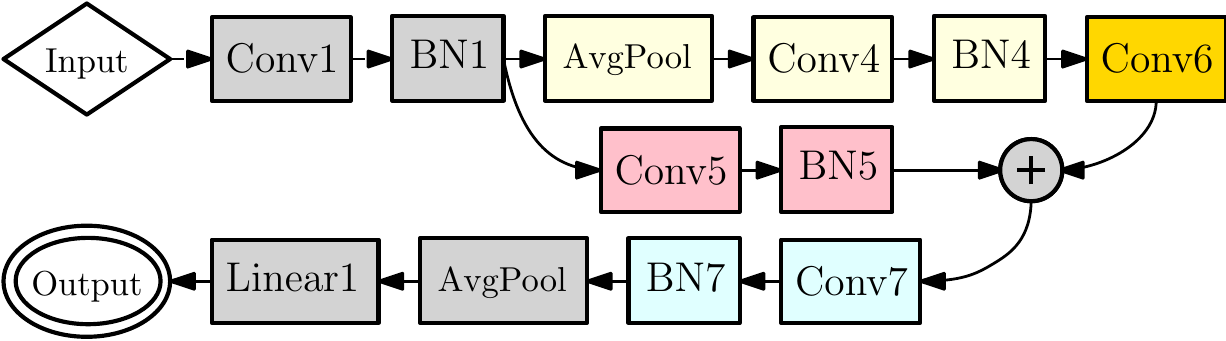}
    	\caption{Constructed sub-network.}\label{fig:constructured_sub_network}
    	\label{fig:1a}		
    \end{subfigure}
    \caption{Redundant removal structures idenfitications and sub-network construction.}
    \label{fig:sub_network_construction}
    \vspace{-5mm}
\end{figure}

We finally construct a sub-network $\mathcal{M}^*$ upon the given DNN $\mathcal{M}$ and the solution $\bm{x}_\hhspg^*$ by \hhspg{}. The solution $\bm{x}_\hhspg^*$ should attain desired target hierarchical group sparsity level and achieve high performance. As illustrated in Figure~\ref{fig:sub_network_construction}, we first traverse the graph to remove the entire vertices and the related edges from $\mathcal{M}$ corresponding to the redundant removal structures being zero, \eg, \texttt{Conv2-BN2}, \texttt{MaxPool-Conv3-BN3} and \texttt{Conv8-BN8} are removed due to $[\bm{x}_\hhspg^*]_{g_2\cup g_3\cup g_8}=\bm{0}$.  Then, we traverse the graph in the second pass to remove the affiliated structures that are dependent on the removed vertices to keep the remaining operations valid, \eg, the first and second columns in $\bm{\widehat{\mathcal{K}}_6}$ are erased since its incoming vertices \texttt{Conv2-BN2} and \texttt{MaxPool-Conv3-BN3} have been removed (see Figure~\ref{fig:zero_groups}). Next, we recursively erase unnecessary vertices and isolated vertices. Isolated vertices refer to the vertices that have neither incoming nor outgoing vertices. Unnecessary vertices refer to the skippable operations, \eg, \texttt{Concat} and \texttt{Add} (between \texttt{Conv7} and \texttt{AvgPool}) become unnecessary. Ultimately, a compact sub-network $\mathcal{M}^*$ is constructed as shown in Figure~\ref{fig:constructured_sub_network}. Fine-tuning the constructed sub-network $\mathcal{M}^*$ is optional and often not necessary, particularly if the removed structures additionally exhibit the zero-invariant property \citep{chen2021oto}. 
\vspace{-2mm}

\section{Numerical Experiments}\label{sec.experiments}
\vspace{-1mm}

In this section, we employ \algacro{} to one-shot automatically train and search within general DNNs to construct compact sub-networks with high performance.  The numerical demonstrations cover extensive DNNs including \demosupnet{} shown in Section~\ref{sec.otov3}, \regnet{}~\citep{radosavovic2020designing}, \stackedunets{}~\citep{ronneberger2015u}, \superresnet{}~\citep{he2016deep,ming_zennas_iccv2021}, and \darts{}~\citep{liu2018darts}, and benchmark datasets, including \cifar{}~\citep{Krizhevsky09}, \fashionmnist{}~\citep{xiao2017online}, \imagenet{}~\citep{deng2009imagenet}, \stl{}~\citep{coates2011analysis} and \svnh{}~\citep{netzer2011reading}. More implementation details of experiments and \algacro{} library and limitations are provided in Appendix~\ref{appendix.implementation_details}. 

\vspace{-3mm}
\begin{table}[h]
    \centering
    \scriptsize
   \caption{\algacro{} on extensive super-networks and datasets.}
   \vspace{-2mm}
    \label{table.otov3_various_networks_datasets}
    \vspace{1mm}
	\resizebox{\linewidth}{!}{
	\begin{tabular}{ cccccc}
	\Xhline{3\arrayrulewidth}
    Backend  & Dataset & Method  & FLOPs (M) & \# of Params (M) & Top-1 Acc. (\%)  \\
    \hline
    \demosupnet{} & \fashionmnist{} & Baseline &  209 & 0.82 & 84.9 \\ 
    \demosupnet{} & \fashionmnist{} & \textbf{\algacro{}} & \textbf{107} & \textbf{0.45} & \textbf{84.7}  \\ 
    \hdashline
    StackedUnets & \svnh{} & Baseline & 184 & 0.80 & 95.3 \\
    StackedUnets & \svnh{} & \textbf{\algacro{}} & \textbf{115} & \textbf{0.37} & \textbf{96.1} \\ 
    \hdashline
    DARTS (8 cells) & STL-10 & Baseline & 614 & 4.05 & 74.6 \\
    DARTS (8 cells) & STL-10 & \textbf{\algacro{}} & \textbf{127} & \textbf{0.64} & \textbf{75.1} \\
    \Xhline{3\arrayrulewidth}
	\end{tabular}
	}
\vspace{-5mm}
\end{table}

\paragraph{\demosupnet{} on \fashionmnist{}.} We first experiment with the \demosupnet{} presented as Figure~\ref{fig:demo_super_network} on \fashionmnist{}. \algacro{} automatically establishes a search space of \demosupnet{} and partitions its trainable variables into a set of groups. \hhspg{} then trains \demosupnet{} from scratch and computes a solution of high performance and hierarchical group-sparsity over the generated search space, which is further utilized to construct a compact sub-network as presented in Figure~\ref{fig:constructured_sub_network}. As shown in Table~\ref{table.otov3_various_networks_datasets}, compared to the super-network, the sub-network utilizes 54\% of parameters and 51\% of FLOPs to achieve a Top-1 validation accuracy 84.7\% which is negligibly lower than the super-network by 0.2\%.

\paragraph{\stackedunets{} on \svnh{}.} We then consider a \stackedunets{} over \svnh{}. The \stackedunets{} is constructed by stacking two standard Unets~\citep{ronneberger2015u} with different down-samplers together, as depicted in Figure~\ref{fig:stacked_unet_trace_graph} in Appendix~\ref{appendix.dependancy_graphs}. We employ \algacro{} to automatically build the search space and train by \hhspg{}. \hhspg{} identifies and projects the redundant structures onto zero and optimizes the remaining important ones to attain excellent performance. As displayed in Figure~\ref{fig:stacked_dependancy_graph}, the right-hand-side Unet is disabled due to \texttt{node-72-node-73-node-74-node-75} identified as redundant.
The path regarding the deepest depth for the left-hand-side Unet, \ie, \texttt{node-13-node-14-node-15-node-19}, is marked as redundant as well. The results by~\algacro{} indicate that the performance gain brought by either composing multiple Unets in parallel or encompassing deeper scaling paths is not significant. \algacro{} also validates the human design since a single Unet with properly selected depths have achieved remarkable success in numerous applications~\citep{ding2022sparsity,weng2019unet}. Furthermore, as presented in Table~\ref{table.otov3_various_networks_datasets}, the sub-network built by~\algacro{} uses 0.37M parameters and 115M FLOPs which is noticeably lighter than the full \stackedunets{} meanwhile significantly outperforms it by 0.8\% in validation accuracy. 

\paragraph{\darts{} (8-Cells) on \stl{}.} We next employ \algacro{} on \darts{} over \stl{}. \darts{} is a complicated network consisting of iteratively stacking multiple cells~\citep{liu2018darts}. Each cell is constructed by spanning a graph wherein every two nodes are connected via multiple operation candidates. \stl{} is an image dataset for the semi-supervising learning, where we conduct the experiments by using its labeled samples. \darts{} has been well explored in the recent years. However, the existing NAS methods studied it based on a \textit{handcrafted} search space beforehand to \textit{locally} pick up one or two important operations to connect every two nodes. We now employ \algacro{} on an eight-cells \darts{} to \textit{automatically} establish its search space, then utilize \hhspg{} to one shot train it and search important structures \textit{globally} as depicted in Figure 6 in the openreview version (due to exceeding maximal dimension in arxiv).  Afterwards, a sub-network is automatically constructed. Quantitatively, the sub-network outperforms the full \darts{} in terms of validation accuracy by 0.5\% by using only about 15\%-20\% of the parameters and the FLOPs of the original network (see Table~\ref{table.otov3_various_networks_datasets}). 
\vspace{-4mm}
\begin{table}[]
\centering
    \vspace{-5mm}
    \caption{\algacro{} over \superresnet{} on \cifar{}.}
	\label{table.superresnet_cifar10}
    \vspace{-1mm}
	\resizebox{\linewidth}{!}{
	\begin{tabular}{lccccc}
	\Xhline{3\arrayrulewidth}
	\multirow{2}{*}{Architecture} & \multirow{2}{*}{Type} & \multirow{2}{*}{Search Space} & \multirow{2}{*}{Top-1 Acc (\%)} & \multirow{2}{*}{\# of Params (M)} & Search Cost \\
    & & & & & (GPU days)\\
	\hline
    Zen-Score-2M~\citep{ming_zennas_iccv2021} & Zero-Shot & ResNet Pool  & 97.5 & 2.0 &  0.5 \\
    TENAS~\citep{chen2021neural} & Zero-Shot & DARTS & 97.4 & 3.8 &  0.04 \\
    SANAS-\darts{}~\citep{hosseini2022saliency} & Gradient & DARTS & 97.5 & 3.2 & \hspace{1.4mm}1.2$^\dagger$ \\
    ISTA-NAS~\citep{he2020milenas} & Gradient & DARTS & 97.5 & 3.3 & 0.1 \\
    CDEP~\citep{rieger2020interpretations} & Gradient & DARTS & 97.2 & 3.2 & \hspace{1.4mm}1.3$^\dagger$ \\
    DARTS (2nd order)~\citep{liu2018darts} & Gradient & DARTS & 97.2 & 3.1 & 1.0 \\
    PrDARTS~\citep{zhou2020theory} & Gradient & DARTS & 97.6 & 3.4 & 0.2\\
    P-DARTS~\citep{chen2019progressive} & Gradient & DARTS & 97.5 & 3.6 & 0.3 \\
    PC-DARTS~\citep{xu2019pc} & Gradient & DARTS & 97.4 & 3.9 &  0.1 \\
	\hdashline
	\textbf{\algacro{}} & Gradient & \superresnet{} & 97.5 & 2.0 & 0.1 \\
	\Xhline{3\arrayrulewidth}
    \multicolumn{4}{l}{The search cost is measured by replicating these methods on an NVIDIA A100 GPU.}\\
    \multicolumn{4}{l}{$^\dagger$ Numbers are approximately scaled based on \citep{hosseini2022saliency}.}
	\end{tabular}
}
 \vspace{-6mm}
\end{table}

\paragraph{\superresnet{} on \cifar{}.} Later on, we switch to a ResNet search space inspired by ZenNAS~\citep{ming_zennas_iccv2021}, referred to as \superresnet{}. ZenNAS~\citep{ming_zennas_iccv2021} uses a ResNet pool to populates massive ResNet candidates and ranks them via zero-shot proxy.  Contraily, we independently construct \superresnet{} by stacking several super-residual blocks with varying depths. Each super-residual blocks contain multiple \texttt{Conv} candidates with kernel sizes as \texttt{3x3}, \texttt{5x5} and \texttt{7x7} separately in parallel (see Figure 7 in the open-review version). \superresnet{} includes the optimal architecture derived from ZenNAS and aims to discover the most suitable sub-networks using H2SPG over the automated generated search space. The sub-network produced by~\algacro{} could reach the benchmark over 97\% validation accuracy. Remark here that \algacro{} and ZenNAS use fewer parameters to achieve competitive performance to the DARTS benchmarks. This is because of the extra data-augmentations such as MixUp~\citep{zhang2017mixup} by ZenNAS, so as \algacro{} to follow the same training settings.

\begin{table}[ht!]
    \centering
    \caption{\algacro{} over \darts{} on \imagenet{} and comparison with state-of-the-art methods.}
    \vspace{-1mm}
	\label{table.darts_imagenet}
	\resizebox{\linewidth}{!}{
	\begin{tabular}{lccccc}
	\Xhline{3\arrayrulewidth}
	\multirow{2}{*}{Architecture} &  \multicolumn{2}{c}{Test Acc. (\%)} & \multirow{2}{*}{\# of Params (M)} & \multirow{2}{*}{FLOPs (M)}  &  \multirow{2}{*}{Search Method}\\
    \cline{2-3} & Top-1 & Top-5 & & \\
	\hline
    Inception-v1~\citep{szegedy2015going} & 69.8 & 89.9 & 6.6 & 1448  & Manual\\
    ShuffleNet 2× (v2)~\citep{ma2018shufflenet} & 74.9 & -- & 5.0 & 591  & Manual\\
    \hdashline
    NASNet-A~\citep{zoph2018learning} & 74.0 & 91.6 & 5.3 & 564  & RL \\
    MnasNet-92~\citep{tan2019mnasnet} & 74.8 & 92.0 & 4.4 & 388  & RL \\
    AmoebaNet-C~\citep{real2019regularized} & 75.7 & 92.4 & 6.4 & 570 & Evolution \\
    \hdashline
	DARTS (2nd order) (\cifar)~\citep{liu2018darts} & 73.3 & 91.3 & 4.7 & 574  & Gradient\\
    P-DARTS (\cifar)~\citep{chen2019progressive} & 75.6 & 92.6 & 4.9 & 557   & Gradient\\
    PC-DARTS (\cifar)~\citep{xu2019pc} & 74.9 & 92.2 & 5.3 & 586  & Gradient\\
    SANAS (\cifar)~\citep{hosseini2022saliency} & 75.2 & 91.7 & -- & --  & Gradient\\
    \hdashline
    ProxylessNAS (\imagenet)~\citep{cai2018proxylessnas} & 75.1 & 92.5 & 7.1 & 465   & Gradient\\
    PC-DARTs (\imagenet)~\citep{xu2019pc} & 75.8 & 92.7 & 5.3 & 597 &  Gradient \\
    ISTA-NAS (\imagenet)~\citep{yang2020ista} & 76.0 & 92.9 & 5.7 & 638  & Gradient \\
    \hdashline
	\textbf{\algacro{}} on \darts{} (\imagenet) & 75.3 & 92.5 & 4.8 & 547 & Gradient\\ 
	\Xhline{3\arrayrulewidth}
    \multicolumn{6}{l}{(\cifar{}) / (\imagenet) refer to using either \cifar{} or \imagenet{} for searching architecture. }.
	\end{tabular}
	}
\vspace{-5mm}
\end{table}

\paragraph{\darts{} (14-Cells) on \imagenet{}.} We now present the benchmark \darts{} network stacked by 14 cells on \imagenet{}. We employ \algacro{} over it to automatically figure out the search space which the code base required specified handcraftness in the past, train by \hhspg{} to figure out redundant structures, and construct a sub-network as depicted in Figure 8 (in the open-review version). Quantitatively, we observe that the sub-network produced by \algacro{} achieves competitive top-1/5 accuracy compared to other state-of-the-arts as presented in Table~\ref{table.darts_imagenet}. Remark here that it is \textit{engineeringly} difficult yet to inject architecture variables and build a multi-level optimization upon a search space being automatically constructed and globally searched. The single-level \hhspg{} does not leverage a validation set and specified auxiliary architecture variables as others to conduct multi-level optimization to favor architecture search and search over the operations without trainable variables, \eg, skip connection. Consequently, our achieved accuracy does not outperform PC-DARTS and ISTA-NAS. We leave further improvement over automated multi-level optimization establishment as future work. 

\paragraph{Ablation Study (\regnet{} on \cifar).} We finally conduct ablation studies over \regnet{}~\citep{radosavovic2020designing} on \cifar{} to demonstrate the necessity and efficacy of hierarchical sparse optimizer \hhspg{} compared to the existing non-hierarchical sparse optimizers, which is the key to the success of \algacro{}. Without loss of generality, we employ \algacro{} over the  RegNet-800M  which has accuracy 95.01\% on \cifar{}, and compare with the latest variant of HSPG, \ie, \dhspg{}~\citep{chen2023otov2}. 
We evaluate them with varying target hierarchical group sparsity levels in problem~(\ref{prob.main}) across a range of $\{0.1, 0.3, 0.5, 0.7, 0.9\}$. As other experiments, \algacro{} automatically constructs its search space, trains via \hhspg{} or \dhspg{}, and establishes the sub-networks without fine-tuning.  The results are from three independent tests under different random seeds, and reported in Table~\ref{table.otov3_regnet_cifar10}.
\vspace{-2mm}
\begin{table}[h]
    \centering
    \scriptsize
   \caption{\algacro{} on \regnet{} on \cifar{}.}
   \vspace{-2mm}
    \label{table.otov3_regnet_cifar10}
\begin{minipage}{.68\textwidth}
    \resizebox{\linewidth}{!}{
	\begin{tabular}{ cccccc}
	\Xhline{3\arrayrulewidth}
    \multirow{2}{*}{Backend}   & \multirow{2}{*}{Method} & \multirow{2}{*}{Optimizer} & Target  & \multirow{2}{*}{\# of Params (M)} & \multirow{2}{*}{Top-1 Acc. (\%)}  \\
    & & & Sparsity  & &  \\
    \hline
      &  &  & 0.1 & 5.56 $\pm$\  0.02  & 95.26 $\pm$\ 0.13 \\
     &  &  & 0.3 & (3.40, \xmark, \xmark)  & (95.01, \xmark, \xmark) \\ 
    RegNet-800M  & \textbf{\algacro{}} & \dhspg{} & 0.5 & (\xmark, \xmark, \xmark) & (\xmark, \xmark, \xmark) \\ 
      &  &  & 0.7 & (\xmark, \xmark, \xmark) & (\xmark, \xmark, \xmark) \\ 
      &  &  & 0.9 & (\xmark, \xmark, \xmark) & (\xmark, \xmark, \xmark) \\
    \hdashline
     & & & 0.1 & 5.58 $\pm$\  0.01  & 95.30 $\pm$\ 0.10 \\
      & &  & 0.3 & 3.54 $\pm$\  0.15 & 95.08 $\pm$\ 0.14 \\ 
    RegNet-800M  & \textbf{\algacro{}} & \textbf{\hhspg{}} & 0.5 & 1.83 $\pm$\ 0.09 & 94.61 $\pm$\ 0.19 \\ 
      &  &  & 0.7 & 1.16 $\pm$\ 0.12 & 91.92 $\pm$\ 0.24 \\ 
     &  &  & 0.9 & 0.82 $\pm$\ 0.17 & 87.91 $\pm$\ 0.32 \\
    \Xhline{3\arrayrulewidth}
	\end{tabular}
	}
\end{minipage}
\hspace{2mm}
\begin{minipage}{.29\textwidth}
    \centering
     \includegraphics[width=\linewidth]{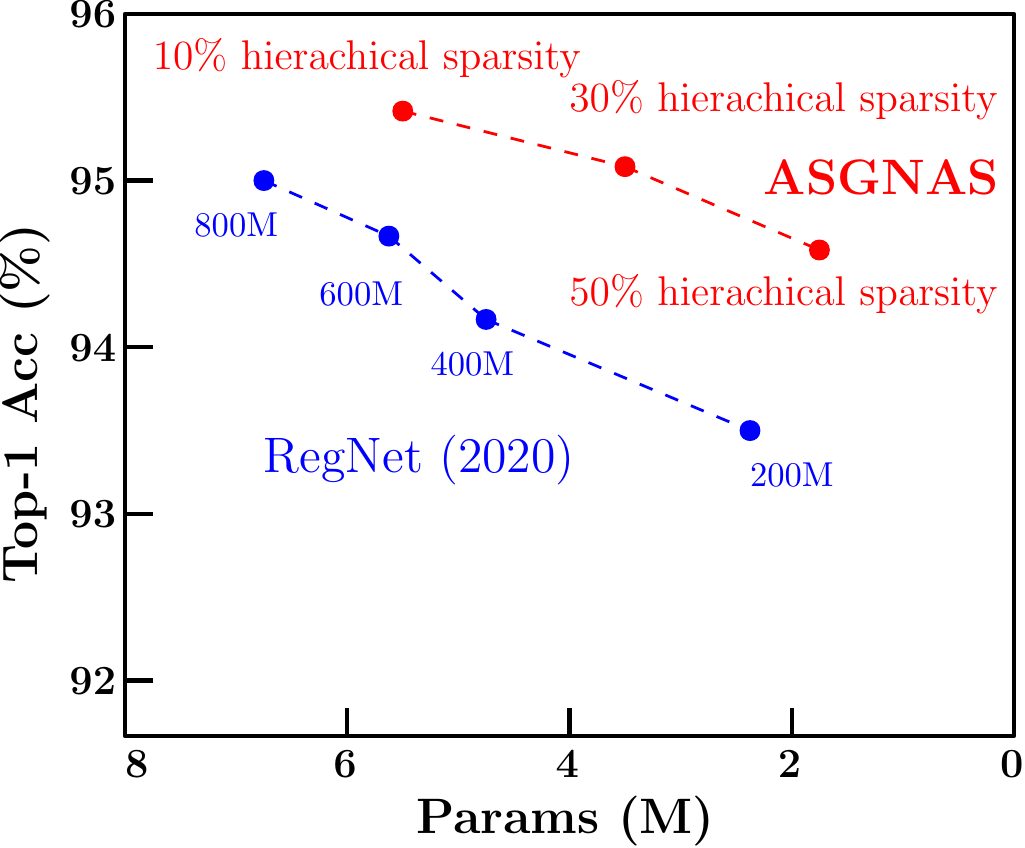}
\end{minipage}
\vspace{-3mm}
\end{table}

    \contour{black}{Sub-networks by~\algacro{} versus Full Networks.}  The sub-networks under varying hierarchical group sparsity levels computed by~\algacro{} with \hhspg{} exhibits the Pareto frontier comparing with the benchmark \regnet{}s. Notably, the sub-networks under sparsity levels of 0.1 and 0.3 outperform the full RegNet-800M. Furthermore, the ones with 0.5 sparsity level outperforms the RegNet(200M-600M), despite utilizes significantly fewer parameters while achieves higher accuracy.

\contour{black}{\hhspg{} versus \dhspg{}.} \dhspg{} often fails when confronts with reasonably large target sparsity levels, denoted by the symbol \xmark. The underlying reason lies in its design, which solely treats problem~(\ref{prob.main}) as an independent and disjoint structured sparsity problem. By disregarding the hierarchy within the network, \dhspg{} easily generates sub-networks that lack validity. Conversely, \hhspg{} takes into account the network hierarchy and successfully addresses the target problem~(\ref{prob.main}).
\vspace{-2mm}

\section{Conclusion}
\vspace{-1mm}

We propose \algacro{}, which is the pioneering automated system to establish search spaces for general DNNs and generates high-performing and compact sub-networks through a novel \hhspg{}. Remarkably, H2SPG stands as the first optimizer to address hierarchical structured sparsity problems for deep learning tasks. \algacro{} significantly minimizes the manual efforts associated with many existing NAS works and pioneers a new trajectory. It also establishes benchmarks regarding automated NAS over general DNNs, which currently requires extensive handcraftness to create search spaces.

\bibliography{iclr2024_conference}
\bibliographystyle{iclr2024_conference}

\newpage
\appendix
\section{Implementation Details}\label{appendix.implementation_details}

We provide more implementation details of \algacro{} library and experiments. 
The official library along with documentations and tutorials will be released to the public after review process.  
\vspace{-2mm}

\subsection{Library Implementations}
\vspace{-2mm}

\paragraph{Overview.} 
Up to the present, the implementation of \algacro{} depends on PyTorch and ONNX (\url{https://onnx.ai}). ONNX is used to obtain the trace graph and the sub-network by modifying the given DNN in ONNX format. \hhspg{} is developed as an instance of the PyTorch optimizer class. 
As a fundamental AI infrastructure, \algacro{} makes a significant breakthrough in AutoML to first enable the search of sub-networks from training general DNNs. Further progress and contributions from both our team and the wider community are necessary to sustain its continued success.
\vspace{-2mm}

\paragraph{Limitations.} The current version of the library relies on ONNX, which means that the DNNs need to be convertible into the ONNX format. Meanwhile, if the given DNN contains unsupported operators, the library may not function normally. To address this, we are committed to maintaining and adding new operators to the library, and leverage contributions from the open-source community in this regard. Additionally, we are actively working on reducing the dependency on ONNX to broaden the library's coverage and compatibility.

Furthermore, for generality, we avoid requiring users to manually introduce auxiliary architecture variables, as seen in the existing gradient-based NAS methods. To search without architecture variables, the current \algacro{} library formulates a single-level hierarchical structured sparsity optimization to identify redundant removal structures based on sparse optimization.  We currently require the removal structures to have trainable variables. Consequently, the operations without trainable variables such as \texttt{skip connection} are not removal for the current version of \algacro{} yet. Identifying and removing operations without trainable variables is an aspect that we consider as future work and plan to address in subsequent updates. 
\vspace{-2mm}

\subsection{Experiment Implementations} 

All experiments were conducted on an NVIDIA A100 GPU. The search cost of \algacro{} was calculated as the runtime of the hierarchical search phase in Algorithm~\ref{alg:main.dhspg}, since it is during this phase that the redundant group candidates are constructed. In our experiments, \hhspg{} follows the existing NAS works~\citep{liu2018darts} by performing 50 epochs for architecture search and evolving the learning rate using a cosine annealing scheduler.


For the \superresnet{} experiments, we adopt the data augmentation technique of MixUp, following the training settings of ZenNAS~\citep{ming_zennas_iccv2021}, and employ a multiple-period cosine annealing scheduler. The maximum number of epochs for the \demosupnet{} and \stackedunets{} is set to 300. In the case of \darts{} on \imagenet{}, we expedite the training process by constructing a sub-network once the desired sparsity level is reached. We then train this sub-network until convergence. All other experiments are carried out in the one-shot manner. 

The initial learning rate is set to 0.1 for most experiments, except for the \darts{} experiments where it is set to 0.01. The lower initial learning rate in \darts{} is due to the absence of auxiliary architecture variables in our \darts{} network, which compute a weighted sum of outputs. Additionally, operations without trainable variables, such as \texttt{skip connections}, are preserved (refer to the limitations). Consequently, the cosine annealing period is repeated twice for the \darts{} experiments to account for the smaller initial learning rate. The mini-batch sizes are selected as 64 for all tested datasets, except for \imagenet{}, where it is set to 128. The target group sparsities are estimated in order to achieve a comparable number of parameters to other benchmarks. This is accomplished by randomly selecting a subset of removal structures to be zero and then calculating the parameter quantities in the constructed remaining sub-networks. 

\section{Complexity Analysis}\label{appendix.time_complexity}

We analyze the time and space complexity in \algacro{} to automatically generate the search space and the hierarchy consideration during \hhspg{} optimization. 

\paragraph{Search Space Construction.} The automatic search space generation (Algorithm~\ref{alg:main.zig_partition}) is primarily a customized graph algorithm designed to identify removal structures and partition trainable variables into a set of hierarchical groups. It contains two main stages: \textit{(i)} establishing the segment graph, and \textit{(ii)} constructing the variable partition. During the first stage, the algorithm traverses the trace graph using a combination of depth-first and breadth-first approaches with specific operations. Consequently, the worst-case time complexity is $\mathcal{O}(|\mathcal{V}|+ |\mathcal{E}|)$ to visit every vertex and edge in the trace graph. The worst-case space complexity equals to $\mathcal{O}(|\mathcal{V}|)$ due to the queue container used in Algorithm~\ref{alg:main.zig_partition} and the cache employed during the recursive depth-first search. In the second stage, the constructed segment graph $(\mathcal{V}_s, \mathcal{E}_s)$ is traversed in a depth-first manner to perform the variable partition. In the worst case scenario, where $(\mathcal{V}_s, \mathcal{E}_s)$ equals $(\mathcal{V}, \mathcal{E})$, the time and space complexities remain the same as $\mathcal{O}(|\mathcal{V}|+ |\mathcal{E}|)$ and $\mathcal{O}(|\mathcal{V}|)$ respectively. In summary, the worst-case time complexity for both stages combined is $\mathcal{O}(|\mathcal{V}|+ |\mathcal{E}|)$, and the worst-case space complexity is $\mathcal{O}(|\mathcal{V}|)$. Therefore, the search space construction can be typically efficiently finished in practice. 

\paragraph{Hierarchy Structured Sparsity Optimization.} Compared to other non-hierarchical optimizers, \hhspg{} takes into account of the hierarchy of the network during optimization to ensure the validity of the generated sub-networks. This is achieved through a hierarchical search phase, which checks if remove one vertex from the segment graph $(\mathcal{V}_s, \mathcal{E}_s)$, determines whether the remaining DNN remains connected from the input to the output. A depth-first search is performed for this purpose, with a worst-case time complexity of $\mathcal{O}(|\mathcal{V}_s|+ |\mathcal{E}_s|)$ and a worst-case space complexity of $\mathcal{O}(|\mathcal{V}_s|)$. Throughout the optimization process, the hierarchy check is only triggered once iteratively over a subset of removal structures (proportional to the target sparsity level). Consequently, the worst-case overall time complexity for the hierarchy check is $\mathcal{O}(|\mathcal{V}_s|^2+ |\mathcal{E}_s|\cdot|\mathcal{V}_s|)$.  The worst-case overall space complexity remains $\mathcal{O}(|\mathcal{V}_s|)$, since the cache used for the hierarchy check is cleaned up after each vertex completes its own check.

It is important to note that although the worst-case time complexity is quadratic in the number of vertices of the constructed segment graph, the hierarchy check can be \textit{efficiently} executed in practice because the number of vertices in the segment graph is typically reasonably limited. Additionally, the hierarchy check only occurs \textit{once} during the entire optimization process, consequently does not bring significant computational overhead to the whole process. 

\section{More Related Works}\label{appendix.more_related_work}

The definition of search space is varying upon different NAS scenarios. In our scenario, we aim to automatically discovering a high-performing compact sub-network given a general DNN. The starting DNN is assumed to cover all operation and connection candidates, and the resulting sub-network serves as its sub-computational-graph. Therefore, the search space of our scenario is defined as a set of removal structures of the given DNN. The sub-network discovery is then formulated as identifying a set of redundant removal structures to construct a valid sub-network. It is noteworthy that our target NAS scenario, definition of search space, and automated generation along with a novel end-to-end automated pipeline address a crucial gap in the NAS realm that has seen rare exploration in the past.

There exists orthogonal search-space (super-network) definitions, along with several works in automation. In the context of \cite{munoz2022automated} and \cite{calhas2022automatic}, the presence of operators in DNNs is preserved, yet their inherent hyperparameters, such as stride and depth for convolutional layers, are searchable. Consequently, the inherent hyperparameters of the existing operators constitute their search space. \cite{zhou2021autospace} defines the search space as the network that encompasses all candidate operations and investigate methods to automatically generate high-quality super-networks that include optimal sub-networks.  Our approach stays complementary and distinct to these definitions and could operate with them jointly.

\section{Graph Visualizations}\label{appendix.dependancy_graphs}
\vspace{-2mm}

We present visualizations generated by the \algacro{} library to provide more intuitive illustrations of the architectures tested in the paper. The visualizations include trace graphs, segment graphs, identified redundant removal structures, and constructed sub-networks. To ensure clear visibility, we highly recommend \textbf{zooming in with an upscale ratio of at least  500\%} to observe finer details and gain a better understanding of the proposed system. \textbf{Some DNNs are omitted to be presented in arxiv version due to exceeding dimension boundary. Please refer to the open-review version.} 

\begin{figure}[H]
    \centering
    \begin{subfigure}[b!]{\textwidth}
    \centering
    \includegraphics[height=0.85\textheight]{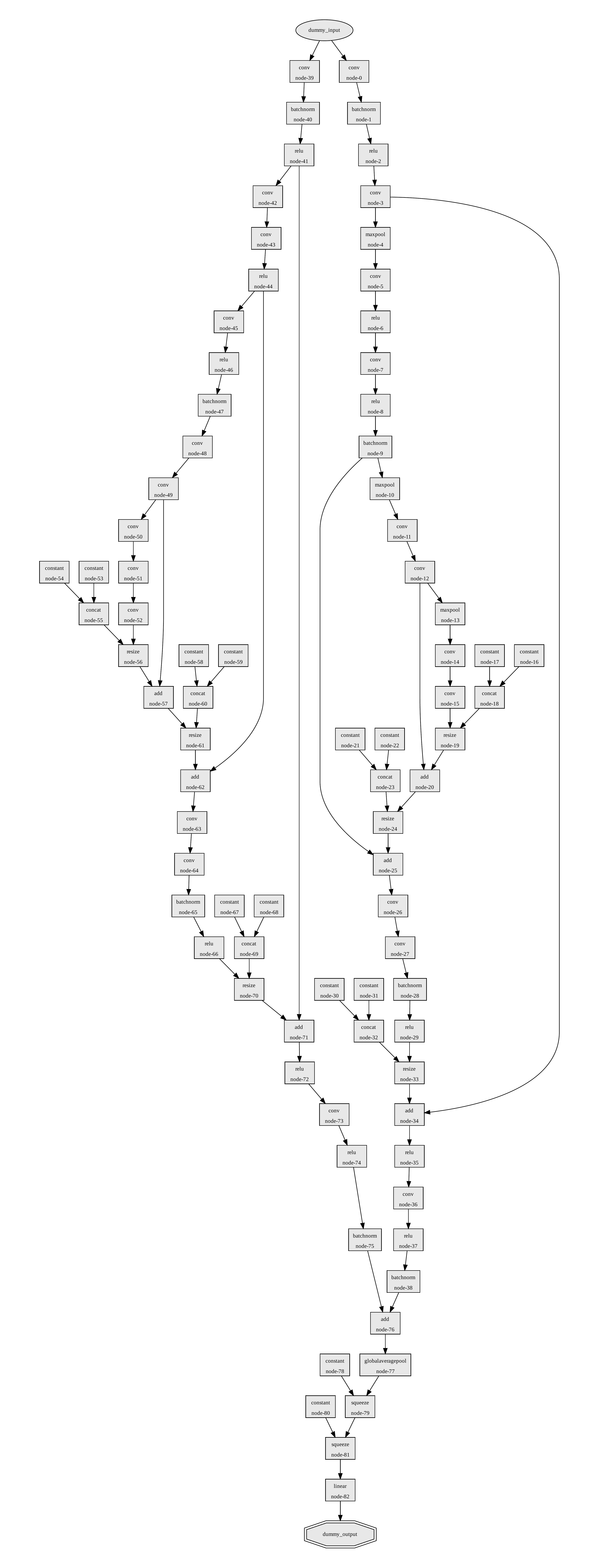}
    \caption{\stackedunets{} trace graph.}
    \label{fig:stacked_unet_trace_graph}
    \end{subfigure}
    \caption{\stackedunets{} illustrations drawn by~\algacro{}.}
    \label{fig:my_label}
\end{figure}
\begin{figure}[H]
\ContinuedFloat
    \begin{subfigure}[b!]{\textwidth}
    \centering
    \includegraphics[height=0.9\textheight]{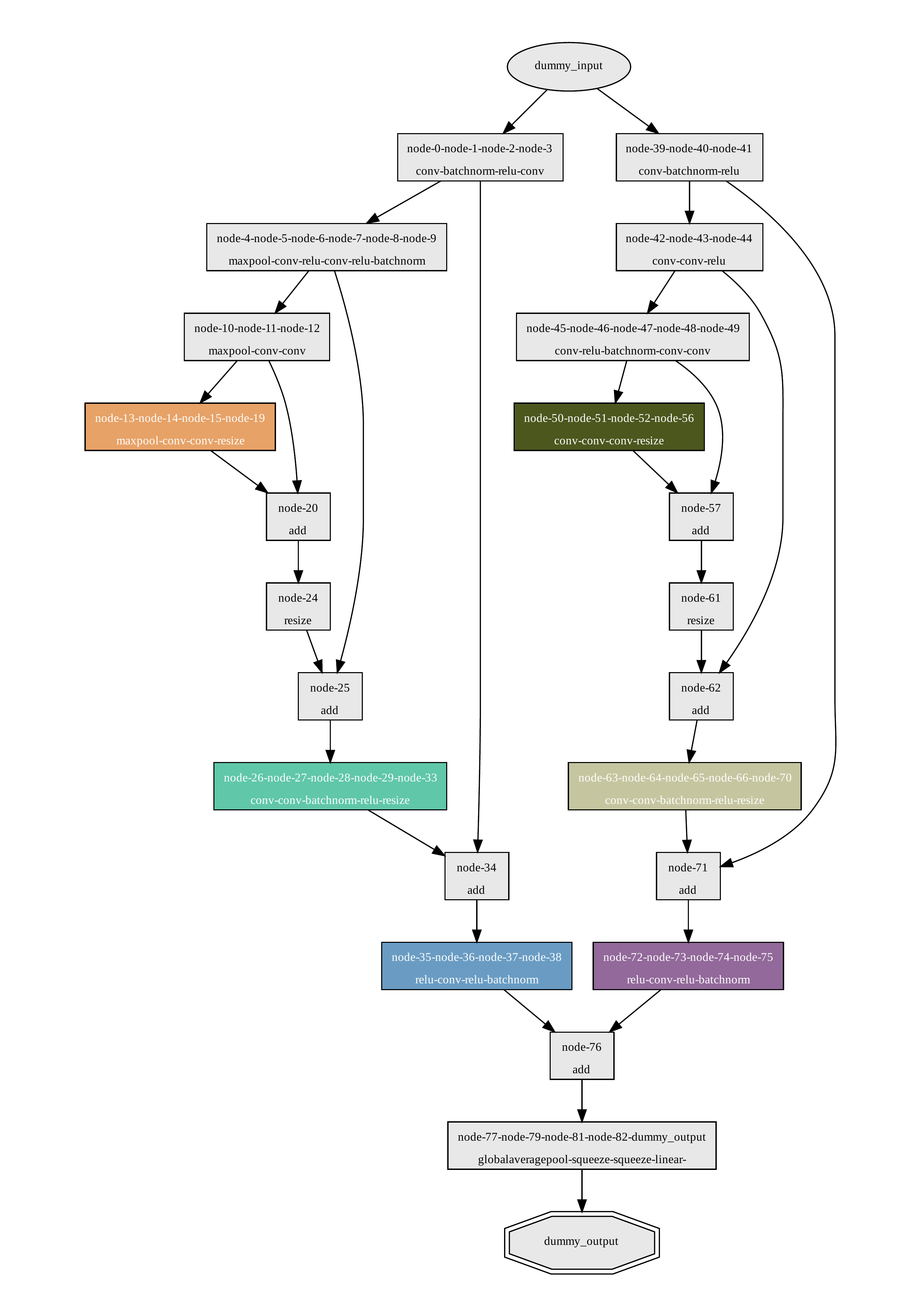}
    \caption{\stackedunets{} segment graph.}
    \label{fig:stacked_dependancy_graph}
    \end{subfigure}
    \caption{\stackedunets{} illustrations drawn by~\algacro{}.}
\end{figure}
\begin{figure}[H]
\ContinuedFloat
    \begin{subfigure}[b!]{\textwidth}
    \centering
    \includegraphics[height=0.9\textheight]{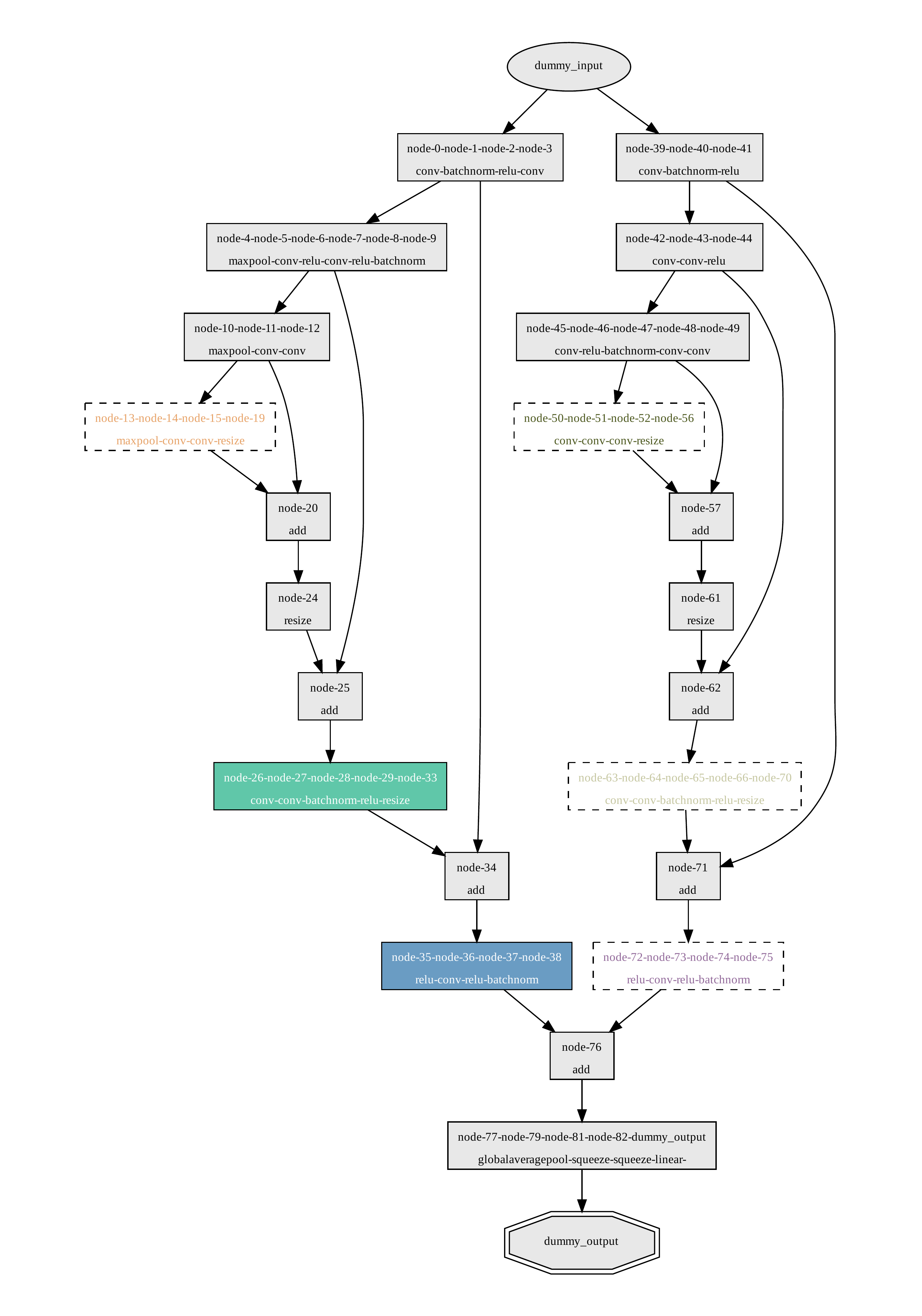}
    \caption{\stackedunets{} segment graph with identified redundant removal structures.}
    \label{fig:stacked_dependancy_graph}
    \end{subfigure}
    \caption{\stackedunets{} illustrations drawn by~\algacro{}.}
\end{figure}
\begin{figure}[H]
\ContinuedFloat
    \begin{subfigure}[b!]{\textwidth}
    \centering
    \includegraphics[height=0.9\textheight]{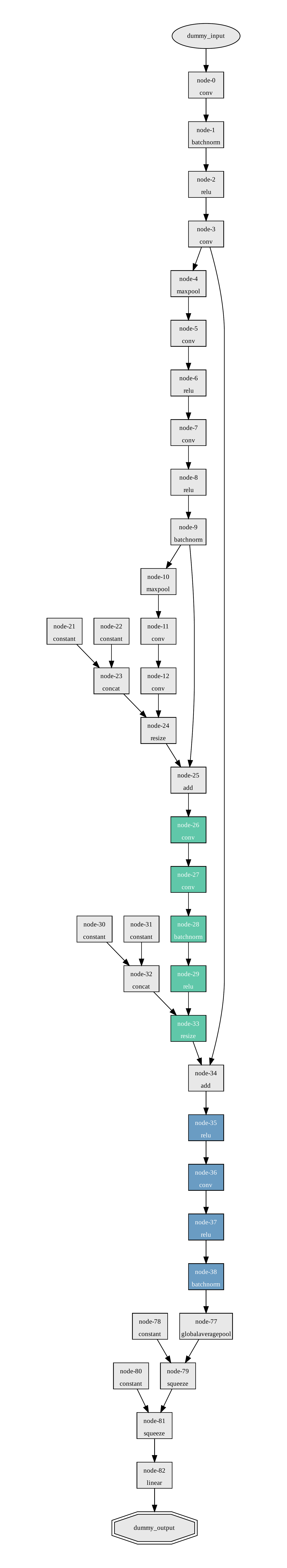}
    \caption{Constructed sub-network upon \stackedunets{}.}
    \label{fig:stacked_subnetwork}
    \end{subfigure}
    \caption{\stackedunets{} illustrations drawn by~\algacro{}.}
\end{figure}

\begin{figure}[H]
    \centering
    \begin{subfigure}[b!]{\textwidth}
    \centering
    \vspace{1mm}
    \includegraphics[height=0.95\textheight]{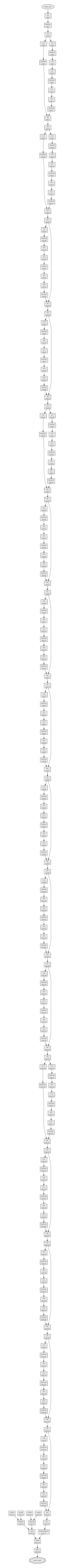}
    \caption{\regnet{} trace graph.}
    \label{fig:regnet_trace_graph}
    \end{subfigure}
    \caption{\regnet{} illustrations drawn by~\algacro{}.}
    \label{fig:my_label}
\end{figure}
\begin{figure}[H]
\ContinuedFloat
    \begin{subfigure}[b!]{\textwidth}
    \centering
    \includegraphics[height=0.95\textheight]{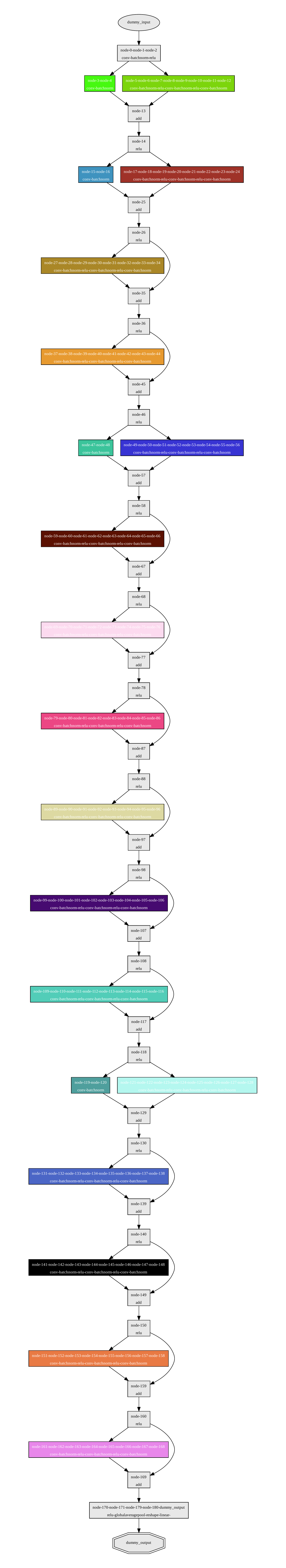}
    \caption{\regnet{} segment graph.}
    \label{fig:regnet_search_space}
    \end{subfigure}
    \caption{\regnet{} illustrations drawn by~\algacro{}.}
\end{figure}
\begin{figure}[H]
\ContinuedFloat
    \begin{subfigure}[b!]{\textwidth}
    \centering
    \includegraphics[height=0.95\textheight]{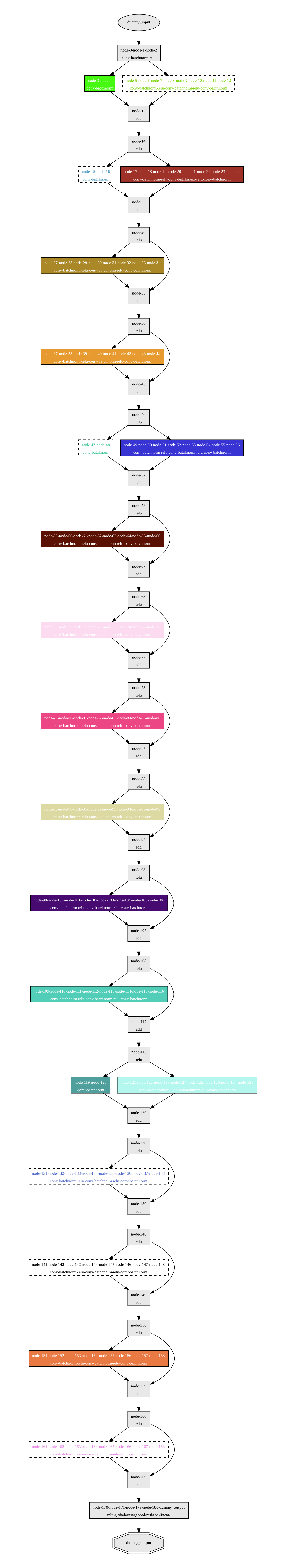}
    \caption{\regnet{} segment graph with identified redundant removal structures.}
    \label{fig:regnet_redundant_structures}
    \end{subfigure}
    \caption{\regnet{} illustrations drawn by~\algacro{}.}
\end{figure}
\begin{figure}[H]
\ContinuedFloat
    \begin{subfigure}[b!]{\textwidth}
    \centering
    \includegraphics[height=0.97\textheight]{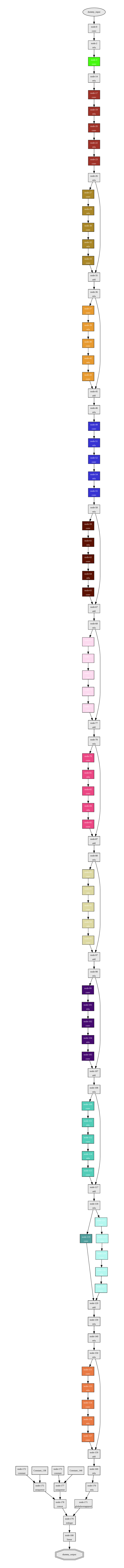}
    \caption{Constructed sub-network upon \regnet{}.}
    \label{fig:regnet_sub_network}
    \end{subfigure}
    \caption{\regnet{} illustrations drawn by~\algacro{}.}
    \label{fig:regnet}
\end{figure}

\end{document}